\documentclass{ecai} 
\usepackage{latexsym}
\usepackage{amssymb}
\usepackage{amsmath}
\usepackage{amsthm}
\usepackage{booktabs}
\usepackage{enumitem}
\usepackage{graphicx}
\usepackage{color}
\usepackage{multirow}

\newcommand{\BibTeX}{B\kern-.05em{\sc i\kern-.025em b}\kern-.08em\TeX}

\begin{document}

%%%%%%%%%%%%%%%%%%%%%%%%%%%%%%%%%%%%%%%%%%%%%%%%%%%%%%%%%%%%%%%%%%%%%%%%

\begin{frontmatter}

%%% Use this command to specify your submission number.
%%% In doubleblind mode, it will be printed on the first page.

\paperid{1985} 

%%% Use this command to specify the title of your paper.

\title{Exploring Boundary-Aware Spatial-Frequency \\
Fusion for Camouflaged Object Detection}
% Exploring Boundary-Aware Spatial-Frequency Fusion for Camouflaged Object Detection
%%% Use this combinations of commands to specify all authors of your 
%%% paper. Use \fnms{} and \snm{} to indicate everyone's first names 
%%% and surname. This will help the publisher with indexing the 
%%% proceedings. Please use a reasonable approximation in case your 
%%% name does not neatly split into "first names" and "surname".
%%% Specifying your ORCID digital identifier is optional. 
%%% Use the \thanks{} command to indicate one or more corresponding 
%%% authors and their email address(es). If so desired, you can specify
%%% author contributions using the \footnote{} command.

\author[A]{\fnms{Song}~\snm{Yu}}
\author[A]{\fnms{Yang}~\snm{Hu}}
\author[A]{\fnms{HaoKang}~\snm{Ding}}
\author[A]{\fnms{ZhiFang}~\snm{Liao}\thanks{Corresponding Author. Email: zfliao@csu.edu.cn.}}
\author[A]{\fnms{YuCheng}~\snm{Song}\thanks{}}

\address[A]{School of Computer Science and Engineering, Central South University, Changsha, China}

%%% Use this environment to include an abstract of your paper.

\begin{abstract}
Camouflaged Object Detection is challenging due to the high degree of similarity between camouflaged objects and their surrounding backgrounds.
Current COD methods mainly rely on edge extraction in the spatial domain and local pixel-level information, neglecting the importance of global structural features.
Additionally, they fail to effectively leverage the importance of phase spectrum information within frequency domain features.
To this end, we propose a COD framework BASFNet based on boundary-aware frequency domain and spatial domain fusion.This method uses dual guided integration of frequency domain and spatial domain features.
A phase-spectrum-based frequency-enhanced edge exploration module (FEEM) and a spatial core segmentation module (SCSM) are introduced to jointly capture the boundary and object features of camouflaged objects.
These features are then effectively integrated through a spatial-frequency fusion interaction module (SFFIM).
Furthermore, the boundary detection is further optimized through an boundary-aware training strategy.
BASFNet outperforms existing state-of-the-art methods on three benchmark datasets, validating the effectiveness of the fusion of frequency and spatial domain information in COD tasks.
\end{abstract}

\end{frontmatter}

%%%%%%%%%%%%%%%%%%%%%%%%%%%%%%%%%%%%%%%%%%%%%%%%%%%%%%%%%%%%%%%%%%%%%%%%

\section{Introduction}

Camouflage refers to a strategy in which organisms alter their appearance to blend in with their surrounding environment, making it visually difficult to detect them.
The research objective of Camouflaged Object Detection (COD) is to identify camouflaged objects that seamlessly integrate with their environment.
This research is of significant value in various fields, including species discovery \cite{perez2012early}, military applications \cite{hwang2024military}, pest monitoring \cite{wu2019ip102}, and polyp segmentation \cite{fan2020pranet}.
In recent years, COD has received increasing attention in the field of computer vision \cite{liang2024systematic}.

With the development of deep learning, several studies \cite{fan2020camouflaged, pang2022zoom, huang2023feature} have utilized deep learning networks to extract multi-level features and perform multi-scale processing on images, significantly improving detection performance.
However, the high similarity between the camouflaged object and the background, especially in boundary recognition, still presents a challenge for accurately detecting camouflaged objects.
To address this issue, several studies \cite{zhu2022can, sun2022boundary, he2023camouflaged, zhai2022mgl, yang2024bi} have introduced edge information in the spatial domain to enhance the detection capabilities of models.
By incorporating edge features, these approaches aim to improve the model's ability to detect camouflaged objects more accurately.
By extracting pixel-level edge features of camouflaged objects, these methods assist the model in more accurately identifying the target. 

\begin{figure}[t]
\centering
\includegraphics[width=\columnwidth]{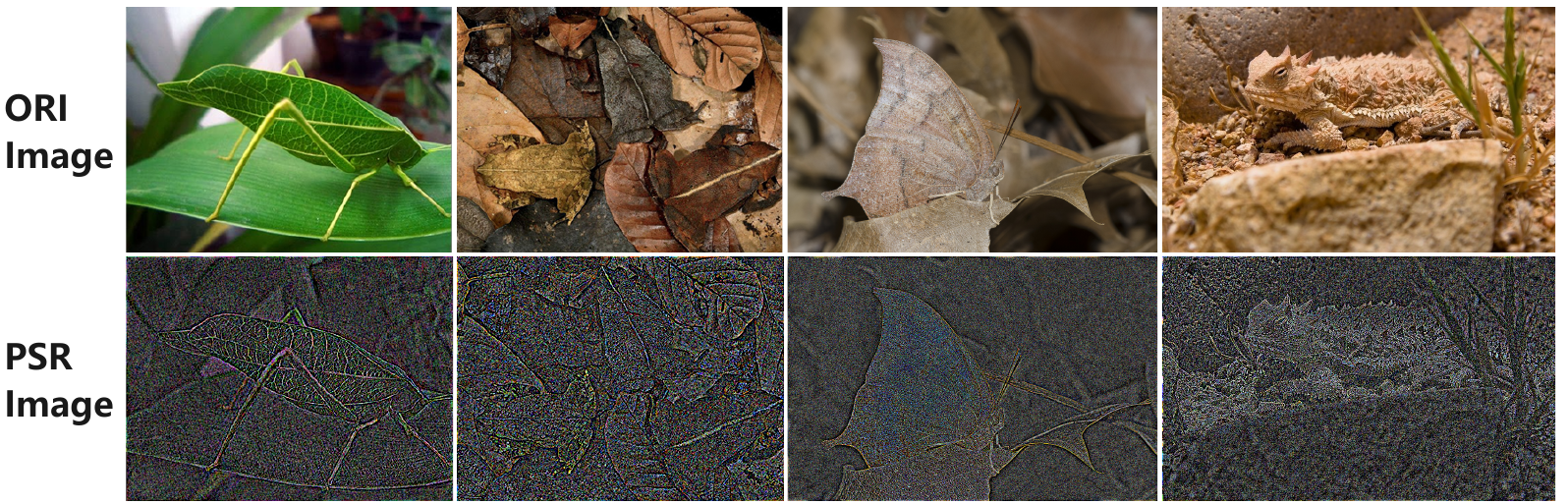}
\caption{Frequency domain analysis of camouflaged images involves frequency decomposition and reconstruction. The first row, labeled "ORI Image," shows the original image, while the second row, "PSR Image," displays the reconstructed image containing only phase information.}
\label{fig1}
\end{figure}

However, the core assumption of these methods is that edge features can precisely distinguish the target from the background.
Most existing COD methods rely on edge information with pixel-level accuracy, but this approach has limitations in the COD task.
First, in complex backgrounds, pixel-level edge features are prone to noise interference.
Relying solely on high-resolution edge features often results in the model being unable to accurately locate the target boundaries.
Secondly, these methods primarily rely on edge extraction in the spatial domain and local pixel-level information, lacking a global perspective.
This local information often fails to effectively capture global structures, leading to imprecise edge information and reduced detection accuracy.
In contrast, frequency domain features, especially phase information, show unique advantages.
As shown in Fig.~\ref{fig1}, the image reconstructed using only phase information can intuitively capture the global structural contours and edge features in the image.
These global structural features can effectively distinguish the target from the complex background, providing clearer structural information for blurred boundaries.
At the same time, they enhance the expression of local features in the spatial domain, thereby compensating for the limitations of traditional spatial-domain methods.
Additionally, by enhancing the boundary characteristics of camouflaged objects, the issue of accurately locating the target boundaries in complex backgrounds can be addressed.
This improvement helps the model better differentiate the target from the surrounding environment.
Therefore, the strategy of combining spatial and frequency domain features, by comprehensively modeling the boundary characteristics of camouflaged objects, can effectively address the limitations of existing methods. 
This approach further enhances the ability to recognize boundaries.

Based on the above discussion, we propose a new framework for COD based on boundary-aware frequency domain and spatial domain fusion, BASFNet.
The model effectively detects camouflaged objects by jointly exploring frequency domain and spatial domain information.
BASFNet consists of three core modules: frequency enhanced edge exploration module (FEEM), spatial core segmentation module (SCSM) and spatial-frequency fusion interaction module (SFFIM).
FEEM captures edge information in the frequency domain by leveraging phase information to emphasize edges and contours, while also exploiting the frequency domain's capability to model global dependencies.
A boundary-aware training strategy is employed to further enhance the edge information, thereby improving edge detection in camouflaged environments.
The SCSM focuses on spatial feature extraction, utilizing a global-guided local learning strategy to effectively distinguish camouflaged objects from complex backgrounds.
Finally, the SFFIM integrates fine edge features from FEEM with segmentation features from SCSM through a dynamic weighting mechanism, ensuring optimal feature fusion and producing a segmentation map with accurate boundaries and complete object recognition.
Our main contributions are as follows:

\begin{figure*}[ht]
\centering
\includegraphics[width=\textwidth]{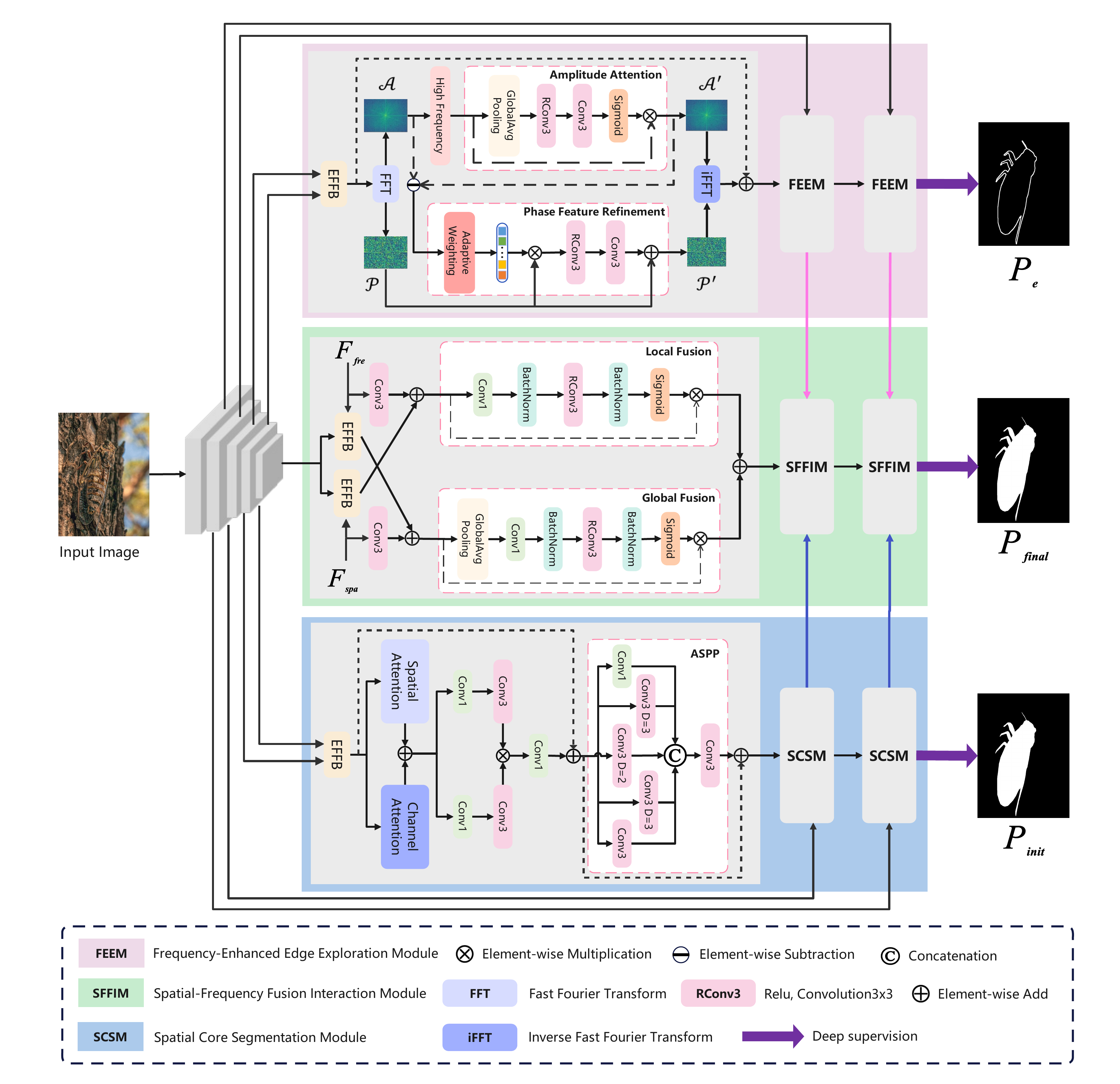}  % 使用 \textwidth 让图片跨越整个双栏
\caption{Overall architecture of the proposed framework BASFNet. The framework consists of three key modules: FEEM, SCSM, and SFFIM, each of which utilizes enhanced feature fusion block (EFFB) for effective feature integration. EFFB enhances the outputs of these modules by leveraging their complementary strengths to boost overall performance. }
\label{fig2}
\end{figure*}

\begin{itemize}
  \item We propose a new frequency and space dual guidance framework based on boundary perception, BASFNet, for camouflaged object detection. The framework achieves target recognition by jointly exploring information in the frequency domain and the spatial domain. It mainly consists of three core modules: FEEM, SCSM and SFFIM.
  \item We propose a boundary-aware training strategy that uses extended edge information to assist the segmentation process, so that the model can better capture the edge features of camouflaged objects when segmenting them.
  \item Extensive experiments demonstrate that our method achieves state-of-the-art (SOTA) performance across most metrics on three benchmark COD datasets.
\end{itemize}

%%%%%%%%%%%%%%%%%%%%%%%%%%%%%%%%%%%%%%%%%%%%%%%%%%%%%%%%%%%%%%%%%%%%%%%%

\section{Related work}

\subsection{Camouflaged Object Detection}

Early camouflaged object detection (COD) methods primarily relied on handcrafted low-level features, such as edges \cite{siricharoen2010robust}, color contrast \cite{huerta2007improving}, geometric cues \cite{tankus2001convexity}, and texture descriptors \cite{bhajantri2006camouflage,kavitha2011efficient}.
However, due to the low contrast between camouflaged objects and their backgrounds, these methods often fail in scenarios involving multiple objects, occlusion, or complex backgrounds.
With the development of deep learning, convolutional neural network (CNN)-based approaches have significantly improved COD performance. Representative works such as SINet \cite{fan2020camouflaged}, ZoomNet \cite{pang2022zoom}, BGNet \cite{xiao2023boundary}, and MGL \cite{zhai2021mutual} have explored strategies including progressive refinement, multi-scale feature fusion, edge-guided supervision, and graph-structured feature modeling, greatly enhancing detection effectiveness.
Recently, the powerful modeling capabilities of Transformer networks \cite{han2022survey,dosovitskiy2020image} have also been introduced into the COD field.
For example, Song et al. \cite{song2023fsnet} proposed a two-stage “Focus-and-Scan” network (FSNet), leveraging the Transformer mechanism to enhance semantic modeling of camouflaged regions. 

Despite these advancements, most methods still primarily rely on local information such as texture and edges in the spatial domain, while neglecting the global structural information embedded in the frequency domain.
To address this, we investigate the phase spectrum in the frequency domain and introduce a boundary-aware spatial-frequency fusion strategy to effectively integrate frequency and spatial features, thereby enhancing the accuracy of camouflaged object detection.

\subsection{Boundary-Aware Feature Learning}

In camouflaged object detection (COD), researchers have adopted the multi-task learning paradigm by incorporating boundary-aware tasks to enhance segmentation accuracy.
Since camouflaged objects are often highly integrated with the background and exhibit vague boundaries, accurately capturing edge information is crucial for precise localization and contour modeling.
Most current COD methods enhance edge perception by incorporating boundary-guided modules or auxiliary edge branches, combined with explicit boundary supervision.
For example, Xiao et al. \cite{xiao2023boundary} proposed the Boundary-guided Context-aware Network (BCNet), which designs a boundary-guided feature interaction module to fuse multi-level features and model the complementarity between object semantics and boundary contours. 
Li et al. \cite{li2024dual} proposed the Boundary and Localization Representation Network (BLRNet), which fuses high-level semantics with positional information to guide low-level features in accurately extracting edges and improving the localization and segmentation of camouflaged objects.

However, most existing methods rely on edge cues extracted from low-level features and focus on enhancing boundary awareness via guidance modules, but they lack thorough modeling of the accuracy, structural integrity, and collaboration between predicted boundaries and target regions.
To address this, we propose a boundary-aware training strategy to enhance boundary recognition, and employ a Spatial-Frequency Fusion Interaction Module (SFFIM) to dynamically integrate spatial and frequency cues, thereby boosting the segmentation performance on camouflaged objects.

\subsection{Frequency Domain Feature Learning}

Frequency-domain feature learning has become an important research direction in visual understanding tasks, especially in scenarios where spatial-domain features fail to provide sufficient information.
By applying the Fourier Transform (FFT) to images or feature maps, frequency-domain methods capture low-frequency components, facilitating the extraction of global contours and structural information that are difficult to obtain in the spatial domain.
In camouflaged object detection, frequency-domain features enhance the model’s ability to recognize subtle structural differences between foreground and background, thus gaining increasing attention.
In recent years, some approaches have attempted to improve COD performance from a frequency perspective. For example, He et al. \cite{he2023camouflaged} proposed the Feature Decomposition and Edge Reconstruction (FEDER) model, which employs learnable wavelets to decompose features into different frequency bands, thereby mitigating the high similarity between foreground and background.
This method further focuses on the most informative frequency bands to extract subtle structural cues that help distinguish camouflaged targets.
More recently, methods such as FSELNet \cite{sun2024frequency} and DSNet \cite{zhang2025frequency} adopt a strategy of fusing frequency-domain and spatial-domain information, leveraging frequency modeling to enhance structural cue representation and combining it with spatial details to achieve more accurate target localization and segmentation.

However, these methods mainly focus on modeling and enhancing frequency-domain information while often overlooking the critical role of the phase spectrum in structure modeling and contour discrimination.
To address this issue, we propose a phase spectrum-based frequency enhancement module (FEEM), which adaptively mines structural boundary information in the frequency domain and integrates it into a boundary-aware COD framework.

%%%%%%%%%%%%%%%%%%%%%%%%%%%%%%%%%%%%%%%%%%%%%%%%%%%%%%%%%%%%%%%%%%%%%%%%

\section{Proposed Method}

\subsection{Framework Architecture}
Camouflaged objects achieve high similarity with their surroundings through adaptive adjustments in color, texture, and shape, making them difficult to distinguish in the spatial domain, especially for methods that rely on local pixel-level modeling.
To address this issue, BASFNet adopts multiple strategies: 1) Unlike previous approaches that mainly focus on frequency amplitude information, BASFNet introduces the phase-spectrum-driven FEEM module, which leverages the phase spectrum’s sensitivity to structural boundaries to effectively extract fine-grained edge cues of camouflaged objects; 2) To enhance the complementary fusion of frequency and spatial features, we design the SFFIM module, which facilitates cross-domain interactive guidance and dynamic integration; 3) During training, BASFNet incorporates a boundary-aware training strategy that applies explicit supervision on boundary regions, effectively improving the accuracy of edge localization.

The overall architecture of BASFNet, shown in Fig.~\ref{fig2}, consists of three key components: FEEM, SCSM and SFFIM. We use the Swin Transformer \cite{liu2021swin} as the backbone network. Given an input image $I$, the network extracts multi-scale features denoted as $X=\left \{ x_{i}, i=1,2,3,4,5  \right \} $, which capture both global and local information of camouflaged objects. To ensure consistency across the different branches while minimizing computational overhead, we apply a $1 \times 1$ convolution to standardize the feature channels to 64. The $x_{1} \rightarrow x_{4}$~features are then fed into the edge frequency detection branch (composed of FEEM modules) and the spatial segmentation branch (composed of SCSM modules) in parallel, extracting edge and object information.The FEEM module emphasizes frequency domain edge features, particularly focusing on phase information, while the SCSM module specializes in local feature learning for spatial segmentation. Finally, the outputs from the FEEM and SCSM modules are passed to the SFFIM module, where the frequency and spatial information are fused and interactively refined to improve the final segmentation. This multi-level fusion strategy significantly enhances the detection and segmentation performance of camouflaged objects in complex backgrounds.

\subsection{Frequency-Enhanced Edge Exploration Module}

Phase conveys more structural information than amplitude and is less susceptible to noise and contrast distortions \cite{oppenheim1981importance}.
In camouflaged object detection (COD), traditional spatial feature extraction methods struggle with the high similarity between camouflaged objects and their background, whereas frequency information enhances edge sensitivity.
The Frequency-Enhanced Edge Exploration Module (FEEM) leverages Fourier transform to extract global frequency features, capturing high-frequency details and improving object boundary visibility.
As shown in Fig.~\ref{fig2}, we use the Enhanced Feature Fusion Block~(EFFB) to integrate these features into an aggregated feature $f_{e}^{1}$.
The Fast Fourier Transform (FFT) is then applied to obtain the amplitude spectrum $\mathcal{A}(f_{e}^{1})$ and phase spectrum $\mathcal{P}(f_{e}^{1})$.
The amplitude, which reflects broader image degradation, plays a key role in recovering detailed phase information and preserving structural details.
We enhance the amplitude features dynamically using a high-frequency enhancement module and an amplitude attention module, leveraging residual amplitude data for phase recovery.
It can be expressed as:
\begin{equation}
\mathcal{A} \left ( f_{e}^{1}  \right ) ,\mathcal{P} \left ( f_{e}^{1}  \right )= \mathcal{F} \left ( f_{e}^{1}  \right )   \label{eq}
\end{equation}
\begin{equation}
{\mathcal{A} }'  \left ( f_{e}^{1}  \right )= \mathcal{H}  \left ( \mathcal{A}\left (f_{e}^{1}   \right )  \right )\cdot \mathcal{A}tt\left ( \mathcal{H}  \left ( \mathcal{A}\left (f_{e}^{1}   \right )  \right ) \right )     \label{eq}
\end{equation}
\noindent where $\mathcal{F} \left ( \cdot \right ) $ represents the fast Fourier transform. $\mathcal{H} \left ( \cdot \right ) $ represents the high-frequency enhancement module, which is a $3\times 3$ convolution with batch normalization operation. $\mathcal{A}tt \left ( \cdot \right ) $ represents the amplitude attention module.

Next, we guide the optimization of phase features through an adaptive weighting module based on amplitude variations, thereby enhancing the global descriptive power of the phase information.
Finally, the enhanced frequency domain features are mapped back to the spatial domain through an Fast Inverse Fourier Transform to produce the final output features:
\begin{equation}
{\mathcal{P} }'  \left ( f_{e}^{1}  \right )= P\left (\mathcal{P} \left ( f_{e}^{1}  \right )\cdot \mathcal{W}\left ( {\mathcal{A} }'  \left ( f_{e}^{1}  \right )- \mathcal{A}  \left ( f_{e}^{1}  \right ) \right )   \right )      \label{eq}
\end{equation}
\begin{equation}
F_{fre}^{i+1}=\mathcal{F} ^{-1}\left ({\mathcal{A} }'  \left ( f_{e}^{1}  \right ),{\mathcal{P} }'  \left ( f_{e}^{1}  \right )  \right ) + f_{e}^{1}  \label{eq}
\end{equation}
\noindent where $\mathcal{W} \left ( \cdot \right ) $is an adaptive weighting module that dynamically adjusts the weight according to the amplitude residual. $P \left ( \cdot \right ) $ is a phase feature Refinement module that further enhances the phase feature. $\mathcal{F} ^{-1} \left ( \cdot \right )$ is an Fast Inverse Fourier Transform.

\begin{figure}[t]
\centering
\includegraphics[width=\columnwidth]{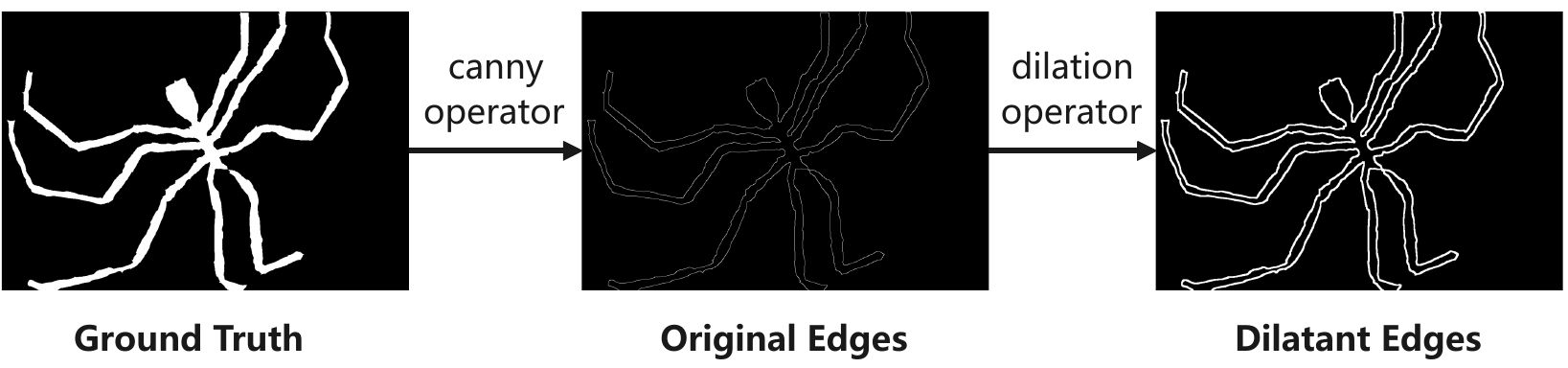}
\caption{The process of producing edge ground truths in boundary-aware enhancement strategy.}
\label{fig3}
\end{figure}

To enhance the boundary awareness capability of the FEEM module, we design a boundary-aware training strategy that combines phase information extraction with expanded edge supervision to assist in boundary learning for camouflaged object segmentation.
As illustrated in Fig.~\ref{fig3}, we first apply the Canny operator \cite{canny1986computational} to the camouflage object mask $G_m$ to generate the initial edge map $G_e$. Then, we perform dilation on $G_e$ using convolution kernels of different sizes $k \times k$ ($ k \in \left \{ {1,3,5,7,9} \right \} $), resulting in the dilated edge map $G_e^d$.
This process broadens the boundary regions and introduces additional transition information between the object and background.
During training, $G_e^d$ serves as the supervision signal for the expanded edges, emphasizing regions with significant edge variations.
This strategy enables the FEEM module to better capture boundary details and improves its edge detection ability under complex backgrounds.
The impact of different kernel sizes on the performance of BSAFNet will be discussed in detail in Section 4.4.

\subsection{Spatial Core Segmentation Module}
As shown in Fig.~\ref{fig2}, the Spatial Core Segmentation Module (SCSM) is designed to focus on object segmentation in the spatial domain and provides high-quality object features for the subsequent Frequency-Spatial Fusion Interaction Module (SFFIM).
First, we use the Enhanced Feature Fusion Block (EFFB) to integrate these features into an aggregated feature $f_{s}^{1}$.
Next, Spatial Attention and Channel Attention mechanisms are applied to highlight important features in the spatial and channel dimensions, respectively.
These attentions adaptively weight salient features, enhancing the segmentation of the camouflaged object’s main region.
The attention features are then summed to produce an enhanced representation.
Subsequently, these features are decomposed into collaborative features through depthwise separable convolutions, establishing non-linear associations: 
\begin{equation}
P^{split}=  DWConv\left (SpaIn\left (SA\left ( f_{s}^{1} \right ) + CA\left ( f_{s}^{1} \right )   \right )  \right )   \label{eq}
\end{equation}
\noindent where $SpaIn \left ( \cdot \right ) $ is the input projection module that extends the input dimension by a factor of two for feature decoupling. 
$DWConv \left ( \cdot \right ) $ is a depth separable convolution for efficient extraction and interaction, $SA \left ( \cdot \right ) $ denotes the Spatial Attention mechanism, $CA \left ( \cdot \right ) $ denotes the Channel Attention mechanism.

The result is then divided into two parts and processed simultaneously through nonlinear activation, and then the result of the co-processing is projected back to the original feature space.
Finally, all enhanced features are pooled and fed into the Atrous Spatial Pyramid Pooling (ASPP) module to capture multi-scale contextual information.
The process is expressed as:
\begin{equation}
f_{s}^{2}, f_{s}^{3}=P^{split}  \label{eq}
\end{equation}
\begin{equation}
f_{s}^{4}=SpaOut\left (GELU\left ( f_{s}^{2}  \right ) \cdot f_{s}^{3}  \right ) +f_{s}^{1}  \label{eq}
\end{equation}
\begin{equation}
F_{spa}^{i+1} = ASPP\left ( f_{s}^{4}  \right )   \label{eq}
\end{equation}
\noindent where $SpaOut \left ( \cdot \right ) $ is the output projection module that restores the dimension to the original feature dimension, and $ASPP \left ( \cdot \right ) $ is the atrous spatial pyramid pooling module that captures multi-scale contextual information.

\subsection{Spatial-Frequency Fusion Interaction Module}
As shown in Fig.~\ref{fig2}, the Spatial-Frequency Fusion Interaction Module (SFFIM) is a core component that aims to deeply fuse and interact features from the FEEM (edge frequency information) and SCSM (spatial segmentation information) modules, optimizing the final segmentation of camouflaged objects.
The SFFIM achieves complementarity between the frequency and spatial domains by efficiently fusing edge and segmentation features.
Multi-level interaction ensures accurate modeling of both boundary and object regions, preserving fine edge details while enhancing object recognition and segmentation.
Initially, the SFFIM module receives the feature outputs from the FEEM, SCSM, and the previous SFFIM module, which are then integrated using the EFFB unit to generate the preliminary fusion features, $Fuse_{f}^{1}$ and $Fuse_{s}^{1}$.
Next, these fused features are input into two parallel processing branches.
One branch uses the importance distribution of local features to enhance fine-grained features through the Local Fusion block to capture local correlations between spatial and frequency features.
The other branch aims to capture the complementary information of spatial and frequency features from a global perspective through the Global Fusion block.
After the features from both branches are fused, all output features are further refined through the feature optimization module.
This additional refinement enhances the quality of the fused features, leading to improved overall accuracy in camouflaged object detection.
The process can be expressed as:
\begin{equation}
Fuse_{s}^{2}=  LF\left [ CBR_{3\times 3 }\left ( F_{spa}^{i}  \right )  +  Fuse_{f}^{1} \right ]  \label{eq}
\end{equation}
\begin{equation}
Fuse_{f}^{2}=  GF\left [ CBR_{3\times 3 }\left ( F_{fre}^{i}  \right )  +  Fuse_{s}^{1} \right ]  \label{eq}
\end{equation}
\begin{equation}
F_{fsue}^{i+1}=Refine\left ( Fuse_{s}^{2} + Fuse_{f}^{2} \right )   \label{eq}
\end{equation}
\noindent where $LF \left ( \cdot \right ) $ represents the local fusion branch, and $GF \left ( \cdot \right ) $ represents the global fusion branch. $Refine \left ( \cdot \right ) $ is a $3\times 3$ convolution and batch normalization operation used to compress channel information and remove redundant features.

\begin{figure}[t]
\centering
\includegraphics[width=0.7\columnwidth]{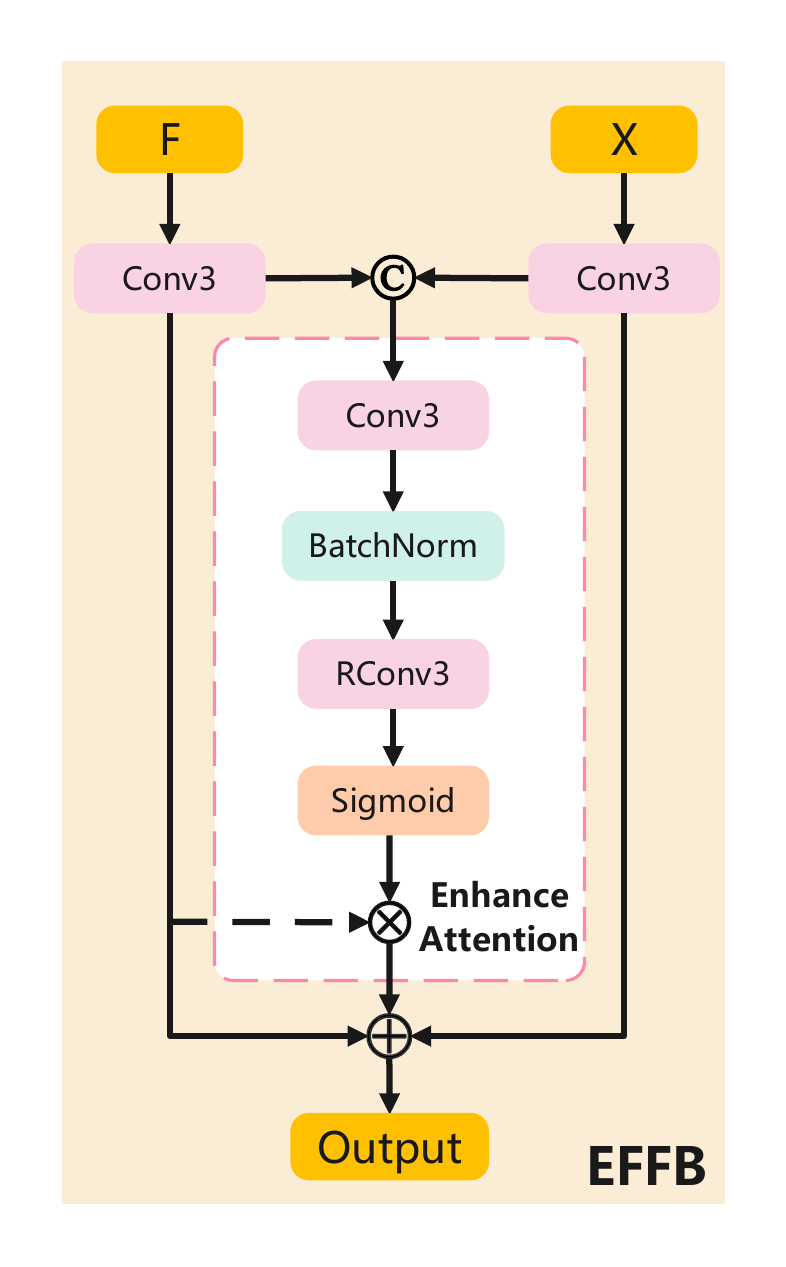}
\caption{Illustration of Enhanced Feature Fusion Block.}
\label{effb}
\end{figure}

\begin{table*}[ht]
\centering
\caption{PERFORMANCE OF DIFFERENT METHODS ON BENCHMARK DATASETS. THE TOP TWO RESULTS ARE HIGHLIGHTED BOLD AND \underline{UNDERLINE},RESPECTIVELY.}
\resizebox{\textwidth}{!}{ % 缩放表格以适应单栏宽度
\begin{tabular}{|l|l|c|c|c|c|c|c|c|c|c|c|c|c|c|c|c|c|}
\hline
\textbf{\multirow{2}{*}{Method}} & \textbf{\multirow{2}{*}{Pub/Years}} & \multicolumn{4}{|c|}{\textbf{CAMO(250 images)}} & \multicolumn{4}{|c|}{\textbf{COD10K(2026 images)}} & \multicolumn{4}{|c|}{\textbf{NC4K(4121 images)}} \\[0.4em]
\cline{3-6} \cline{7-10} \cline{11-14} \cline{15-18}
 &  & \textbf{$S_{m}\uparrow$} & \textbf{$E_{m}\uparrow$} & \textbf{$F_{\beta }^{\omega}\uparrow$} & \textbf{$\mathcal{M}\downarrow $} & \textbf{$S_{m}\uparrow$} & \textbf{$E_{m}\uparrow$} & \textbf{$F_{\beta }^{\omega}\uparrow$} & \textbf{$\mathcal{M}\downarrow $} & \textbf{$S_{m}\uparrow$} & \textbf{$E_{m}\uparrow$} & \textbf{$F_{\beta }^{\omega}\uparrow$} & \textbf{$\mathcal{M}\downarrow $} \\[0.4em]
\hline
SINet-v2 \cite{fan2021concealed} & TPAMI'2021 & 0.820 & 0.882 & 0.743 & 0.070 & 0.815 & 0.887 & 0.680 & 0.037 & 0.840 & 0.903 & 0.805 & 0.048 \\
ZoomNet \cite{pang2022zoom} & CVPR'2022 & 0.820 & 0.892 & 0.752 & 0.066 & 0.838 & 0.911 & 0.729 & 0.029 & 0.853 & 0.912 & 0.784 & 0.043 \\
BSANet \cite{zhu2022can} & AAAI'2022 & 0.796 & 0.851 & 0.717 & 0.079 & 0.818 & 0.891 & 0.699 & 0.034 & 0.841 & 0.903 & 0.778 & 0.048 \\
BGNet \cite{sun2022boundary} & IJCAI'2022 & 0.812 & 0.870 & 0.749 & 0.073 & 0.831 & 0.901 & 0.722 & 0.033 & 0.851 & 0.907 & 0.788 & 0.044 \\
FEDER \cite{he2023camouflaged} & CVPR'2023 & 0.804 & 0.867 & 0.806 & 0.071 & 0.822 & 0.900 & 0.751 & 0.032 & 0.847 & 0.907 & 0.824 & 0.044 \\
MGL \cite{zhai2022mgl} & TIP'2023 & 0.774 & 0.848 & 0.684 & 0.085 & 0.813 & 0.882 & 0.682 & 0.034 & 0.831 & 0.892 & 0.751 & 0.051 \\
FPNet \cite{cong2023frequency} & ACMMM'2023 & 0.852 & 0.905 & 0.806 & 0.056 & 0.850 & 0.913 & 0.748 & 0.029 & - & - & - & - \\
FSPNet \cite{huang2023feature} & CVPR'2023 & 0.856 & 0.899 & 0.799 & 0.050 & 0.851 & 0.895 & 0.735 & 0.026 & 0.879 & 0.915 & 0.816 & 0.035 \\
LGPNet\cite{tong2024local} & CVIU'2024 & 0.817 & 0.873 & - & 0.070 & 0.832 & 0.899 & - & 0.032 & 0.856 & 0.908 & - & 0.044 \\
HAITNet \cite{phung2024hierarchically} & ICME'2024 & 0.859 & 0.925 & \textbf{0.850} & 0.049 & 0.851 & 0.922 & \textbf{0.793} & 0.024 & 0.879 & 0.932 & \textbf{0.858} & 0.033 \\
BIRNet \cite{yang2024bi} & ICME'2024 & 0.870 & 0.926 & 0.829 & \underline{0.044} & 0.862 & 0.928 & 0.773 & 0.024 & \underline{0.886} & 0.936 & 0.841 & 0.032 \\
CamoFormer \cite{yin2024camoformer} & TPAMI'2024 & \underline{0.872} & \underline{0.929} & \underline{0.831} & 0.046 & \underline{0.869} & \underline{0.932} & 0.786 & \underline{0.023} & \textbf{0.892} & \underline{0.939} & 0.847 & \underline{0.030} \\
DSAM \cite{yu2024exploring} & ACMMM'2024 & 0.832 & 0.913 & 0.794 & 0.061 & 0.846 & 0.921 & 0.760 & 0.033 & 0.871 & 0.932 & 0.826 & 0.040 \\
DSNet \cite{yang2025camouflaged} & PR'2025 & 0.817 & 0.870 & 0.726 & 0.073 & 0.809 & 0.878 & 0.657 & 0.038 & 0.843 & 0.894 & 0.753 & 0.050 \\
BSAFNet(OUR) &  & \textbf{0.884} & \textbf{0.939} & \textbf{0.850} & \textbf{0.037} & \textbf{0.871} & \textbf{0.941} & \underline{0.790} & \textbf{0.022} & \textbf{0.892} & \textbf{0.944} & \underline{0.850} & \textbf{0.029} \\
\hline
\end{tabular}
}
\label{tab:methods_comparison}
\end{table*}

\begin{figure*}[ht]
\centering
\includegraphics[width=\textwidth]{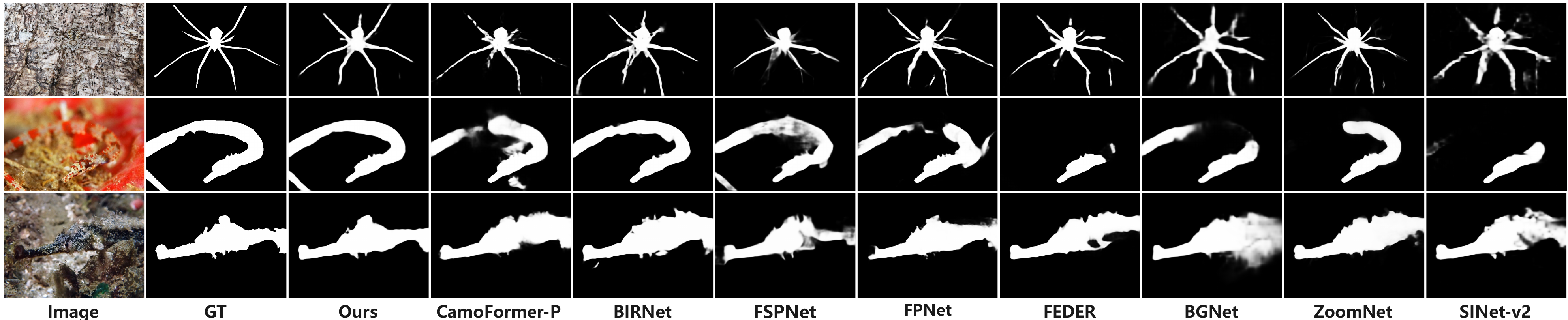}  % 使用 \textwidth 让图片跨越整个双栏
\caption{Visual comparisons of our BIRNet and the competing methods.}
\label{fig4}
\end{figure*}

\subsection{Enhanced Feature Fusion Block}
As shown in Fig.~\ref{effb}, the Enhanced Feature Fusion Block (EFFB) is a module designed to integrate and refine multi-source heterogeneous features, aiming to enhance the collaborative representation capability of different modules such as FEEM, SCSM, and SFFIM.
EFFB extracts rich contextual features through parallel convolution operations and employs an Enhance Attention module to guide the network's focus toward more discriminative semantic regions.
It selectively strengthens key feature channels, thereby improving the semantic completeness and discriminative power of the fused features.
Specifically, EFFB first concatenates the feature map $F_{fs}^{1}$ from the FEEM or SCSM module with the feature map $F_{fuse}^{1}$ from the SFFIM module to fuse their spatial structures and semantic representations.
Then, the Enhance Attention mechanism is applied to generate the attention weights required for targeted enhancement.
The fused features are subsequently further refined by integrating salient guidance features from FEEM or SCSM, ultimately producing feature maps with more complete semantic representation and finer boundary perception. This process can be formalized as:
\begin{equation}
F_{enhance} = Enhance\mathcal{A}tt\left ( Cat\left ( F_{fs}^{1}, F_{fuse}^{1} \right )  \right ) \cdot F_{fs}^{1} \label{eq}
\end{equation}
\begin{equation}
Fuse_{fs}^{i+1} = F_{enhance} + Fuse_{fs}^{1} + F_{fuse}^{1}   \label{eq}
\end{equation}
\noindent where $Cat\left ( \cdot ,\cdot  \right ) $ denotes the concatenation operation, and $Enhance\mathcal{A}tt\left ( \cdot  \right )$ represents the Enhance Attention module, a channel attention mechanism composed of a $3\times 3$ convolution, Batch Normalization, and a Sigmoid activation function.

\subsection{Loss Functions}
To optimize the performance of BASFNet in camouflaged object detection tasks, we design a multi-branch hybrid loss function that balances the requirements of edge detection and object segmentation, fully utilizing the complementarity of frequency and spatial domain features.
The edge prediction $P_{e}$ is optimized using a weighted BCE loss to emphasize the importance of accurate edge detection, especially in challenging areas where the object boundaries are unclear or closely resemble the background. 
For the initial prediction $P_{init}$ and final prediction $P_{final}$, we apply the weighted binary cross-entropy loss ($\mathcal{L}_{BCE}^{w}$) and weighted IOU loss ($\mathcal{L}_{IOU}^{w}$) to handle class imbalance and improve segmentation accuracy:
\begin{equation}
\mathcal{L}_{seg}^{init}=\mathcal{L}_{BCE}^{w}\left ( P_{init},G_{m} \right ) + \mathcal{L}_{IOU}^{w}\left ( P_{init},G_{m} \right )   \label{eq}
\end{equation}
\begin{equation}
\mathcal{L}_{seg}^{final}=\mathcal{L}_{BCE}^{w}\left ( P_{final},G_{m} \right ) + \mathcal{L}_{IOU}^{w}\left ( P_{final},G_{m} \right )   \label{eq}
\end{equation}
\noindent where $G_{m}$ is the ground truth mask for the camouflaged object. Finally, the overall loss function for BASFNet is:
\begin{equation}
\mathcal{L}_{total}= \mathcal{L}_{seg}^{init} + \mathcal{L}_{seg}^{final} + \gamma \cdot \mathcal{L}_{BCE}^{w}\left ( P_{e},G_{e}^{d} \right )   \label{eq}
\end{equation}
\noindent where $\gamma$ is a weight term that controls the importance of the final prediction in the segmentation task. $G_{e}^{d}$ denotes the dilated version of the original edge map $G_{e}$. This operation expands the edge regions, enhancing their saliency and spatial coverage, which facilitates more effective boundary guidance in subsequent feature processing.

\begin{table*}[t]
\caption{ABLATION EXPERIMENTS FOR DIFFERENT KEY COMPONENTS OF BASFNet.}
\begin{center}
\renewcommand{\arraystretch}{1.2} % 调整整体行高
% \resizebox{\columnwidth}{!}{% 调整到单栏宽度
\resizebox{0.7\textwidth}{!}{ % 缩放表格以适应单栏宽度
\begin{tabular}{|c|c|c|c|c|c|c|c|c|c|c|c|c|}
\hline
\textbf{\multirow{2}{*}{Baseline}} & \textbf{\multirow{2}{*}{SCSM}} & \textbf{\multirow{2}{*}{FEEM}} & \textbf{\multirow{2}{*}{SFFIM}} & \textbf{\multirow{2}{*}{EFFB}} & \multicolumn{4}{|c|}{\textbf{COD10K(2026 images)}}  & \multicolumn{4}{|c|}{\textbf{NC4K(4121 images)}} \\
\cline{6-9} \cline{10-13}
 & & & &  & \textbf{$S_{m}\uparrow$} & \textbf{$E_{m}\uparrow$} & \textbf{$F_{\beta }^{\omega}\uparrow$} & \textbf{$\mathcal{M}\downarrow $} & \textbf{$S_{m}\uparrow$} & \textbf{$E_{m}\uparrow$} & \textbf{$F_{\beta }^{\omega}\uparrow$} & \textbf{$\mathcal{M}\downarrow $} \\
\hline
$\surd $ & & & & & 0.662 & 0.702 & 0.310 & 0.139 & 0.741 & 0.748 & 0.446 & 0.139 \\
$\surd $ & $\surd $ & & & & 0.853 & 0.915 & 0.751 & 0.026 & 0.875 & 0.922 & 0.818 & 0.035\\
$\surd $ & $\surd $ & $\surd $ & & & 0.863 & 0.928 & 0.763 & 0.024 & 0.888 & 0.936 & 0.832 & 0.032\\
$\surd $ & $\surd $ & & $\surd $ & & 0.856 & 0.925 & 0.759 & 0.025 & 0.873 & 0.933 & 0.823 & 0.034\\
$\surd $ & & $\surd $& $\surd $ & & 0.863 & 0.933 & 0.768 & 0.024 & 0.887 & 0.938 & 0.835 & 0.032\\
$\surd $ & $\surd $ & $\surd $ & $\surd $ & & 0.865 & 0.935 & 0.776 & 0.023 & 0.889 & 0.942 & 0.843 & 0.030\\
$\surd $ & $\surd $ & $\surd $ & $\surd $ & $\surd $ & \textbf{0.871} & \textbf{0.941} & \textbf{0.790} & \textbf{0.022} & \textbf{0.892} & \textbf{0.944} & \textbf{0.850} & \textbf{0.029}\\
\hline
\end{tabular}
} % 结束 resizebox
\label{tab2}
\end{center}
\end{table*}

%%%%%%%%%%%%%%%%%%%%%%%%%%%%%%%%%%%%%%%%%%%%%%%%%%%%%%%%%%%%%%%%%%%%%%%%

\section{Experiments}

\subsection{Experiment Setup}
\noindent\textbf{Datasets.}
We used three public benchmark datasets for performance evaluation: CAMO \cite{le2019anabranch}, COD10K \cite{fan2020camouflaged}, and NC4K \cite{lv2021simultaneously}.
The CAMO dataset contains 1250 images of camouflaged objects, 1000 of which are used for training and the remaining 250 for testing.
The COD10K dataset contains 5066 images of camouflaged objects, of which 3040 are used for training and 2026 are used for testing.
NC4K is a larger COD test dataset, which contains 4121 camouflaged object images from the Internet.

\noindent\textbf{Evalutaion Criteria.}
We used four widely used metrics to evaluate the performance, including Mean Absolute Error ($\mathcal{M} $) \cite{perazzi2012saliency}, weighted F-measure ($F_{\beta }^{\omega }$) \cite{margolin2014evaluate}, S-measure ($S_{m}$) \cite{fan2017structure}, and mean E-measure ($E_{m}$) \cite{fan2018enhanced}.

\subsection{Implementation  Details}
The proposed BASFNet is implemented in Pytorch on an NVIDIA GeForce RTX 4090.
The pre-trained Swin Transformer model is used as the backbone network, with a learning rate of 1e-4, following a linear warm-up and decay strategy (decayed by a factor of 10 every 50 cycles).
For data augmentation, random horizontal flipping is applied, and all training images are resized to $384 \times 384$.
The model is trained for 100 epochs with a batch size of 8.

\subsection{Comparison with state-of-the-art methods}
To validate the effectiveness of BASFNet, we compare it with 14 representative camouflaged object detection methods, including SINet-v2 \cite{fan2021concealed}, ZoomNet \cite{pang2022zoom}, BSANet \cite{zhu2022can}, BGNet \cite{sun2022boundary}, FEDER \cite{he2023camouflaged}, MGL \cite{zhai2022mgl}, FPNet \cite{cong2023frequency}, FSPNet \cite{huang2023feature}, LGPNet \cite{tong2024local}, HAITNet \cite{phung2024hierarchically}, BIRNet \cite{yang2024bi}, CamoFormer \cite{yin2024camoformer}, DSAM \cite{yu2024exploring}, and DSNet \cite{yang2025camouflaged}.
To ensure fairness of the comparison, all prediction results are from the pre-trained models provided by the authors or generated through available source code.

\noindent\textbf{Quantitative comparison.}
As shown in Table\ref{tab:methods_comparison}, we compare the quantitative results of BASFNet with 14 state-of-the-art methods on the CAMO, COD10K, and NC4K datasets. The results demonstrate that BASFNet consistently outperforms all other methods across the four evaluation metrics, proving its superiority and robustness across different datasets. On the CAMO dataset, BASFNet achieves an average improvement of 1.36$\%$, 1.10$\%$, 2.24$\%$, and 24.32$\%$ over the second-best method CamoFormer \cite{yin2024camoformer} in terms of $S_{m}$, $E_{m}$, $F_{\beta}^{\omega}$, and MAE, respectively.
Compared to BIRNet \cite{yang2024bi}, which incorporates boundary information to assist segmentation, our method improves $S_{m}$, $E_{m}$, and $F_{\beta}^{\omega}$ by 1.61$\%$, 2.98$\%$, and 4.46$\%$, respectively, and reduces $MAE$ by 31.81$\%$ on the CAMO dataset.
Furthermore, compared to the frequency-based FPNet \cite{cong2023frequency}, BASFNet achieves average improvements of 2.41$\%$, 1.40$\%$, and 2.53$\%$ in $S_{m}$, $E_{m}$, and $F_{\beta}^{\omega}$, respectively, and reduces MAE by 15.91$\%$ on the COD10K \cite{fan2020camouflaged} dataset.
This performance gain benefits from the complementary enhancements of FEEM, SCSM, and SFFIM in both frequency and spatial domains, which jointly optimize semantic representation and boundary modeling.

\noindent\textbf{Visual comparison.}
Fig.~\ref{fig4} presents a visual comparison of our approach with several representative COD methods.
When tested in complex backgrounds, BASFNet performed better than other COD methods.
For example, in the second row, our method successfully detects the entire camouflaged object boundary and interior region, while other models fail to fully recognize the target.
This result highlights our model's ability to accurately capture target details in complex backgrounds, as well as its significant advantage in boundary refinement.

\begin{table}[t]
\caption{ABLAITON STUDY OF EDGE-AWARE TRAINING STRATEGY}
\begin{center}
\renewcommand{\arraystretch}{1.1} % 调整整体行高
\resizebox{\columnwidth}{!}{% 调整到单栏宽度
\begin{tabular}{|c|c|c|c|c|c|c|c|c|}
\hline
\textbf{\multirow{2}{*}{Kernel Size($k$)}} &  \multicolumn{4}{|c|}{\textbf{COD10K(2026 images)}}  & \multicolumn{4}{|c|}{\textbf{NC4K(4121 images)}} \\
\cline{2-5} \cline{6-9}
 & \textbf{$S_{m}\uparrow$} & \textbf{$E_{m}\uparrow$} & \textbf{$F_{\beta }^{\omega}\uparrow$} & \textbf{$\mathcal{M}\downarrow $} & \textbf{$S_{m}\uparrow$} & \textbf{$E_{m}\uparrow$} & \textbf{$F_{\beta }^{\omega}\uparrow$} & \textbf{$\mathcal{M}\downarrow $} \\
\hline
$k$ = 1 & 0.868 & 0.939 & 0.783 & 0.023 & 0.891 & 0.942 & 0.845 & 0.031 \\
$k$ = 3 & 0.870 & \textbf{0.941} & 0.788 & \textbf{0.022} & 0.891 & 0.943 & 0.849 & 0.030 \\
$k$ = 5 & \textbf{0.871} & \textbf{0.941} & \textbf{0.790} & \textbf{0.022} & \textbf{0.892} & \textbf{0.944} & \textbf{0.850} & \textbf{0.029}\\
$k$ = 7 & 0.862 & 0.933 & 0.771 & 0.024 & 0.889 & 0.940 & 0.841 & 0.031 \\
$k$ = 9 & 0.861 & 0.935 & 0.772 & 0.024 & 0.887 & 0.940 & 0.839 & 0.032 \\
\hline
\end{tabular}
} % 结束 resizebox
\label{tab3}
\end{center}
\end{table}

\subsection{Ablation study}
To analyze the individual impact of each module and evaluate its contribution to the overall performance of the model, we conducted ablation experiments on the two largest datasets, COD10K \cite{fan2020camouflaged} and NC4K \cite{lv2021simultaneously}, to evaluate the impact of different modules on the overall performance.

\noindent\textbf{Effectiveness of Model Components.}
We conducted an ablation study on the components of BASFNet, and the results are summarized in Table \ref{tab2}. The model with SCSM, FEEM, SFFIM, and EFFB modules removed from BASFNet is used as the baseline. After introducing the SCSM module into the baseline, the model's $S_{m}$ on COD10K significantly increased from 0.662 to 0.853, an improvement of 22.39$\%$; $E_{m}$ improved by 23.27$\%$; $F_{\beta}^{\omega}$ increased dramatically by 85.35$\%$; and $\mathcal{M}$ dropped to 0.026, indicating more accurate detection of camouflaged objects with clearer boundaries.
With the further addition of the FEEM module, the performance on the NC4K dataset improved, with $F_{\beta}^{\omega}$ increasing from 0.818 to 0.832, demonstrating that frequency-guided enhancement can further optimize the perception of edges and salient regions. The SFFIM module brought the most significant improvement in boundary detection ability: on NC4K, $E_{m}$ rose from 0.936 to 0.942, and MAE further decreased to 0.030, highlighting the importance of the spatial-frequency fusion module in modeling camouflaged edge information.
Finally, the addition of the EFFB module resulted in optimal performance across all metrics. On COD10K, $F_{\beta}^{\omega}$ reached 0.790, a 1.77$\%$ improvement over the configuration without this module, while MAE further dropped to 0.022, indicating finer reconstruction of edge details. Similarly, the performance on NC4K steadily improved, with $S_{m}$ reaching 0.892, $F_{\beta}^{\omega}$ rising to 0.850, and MAE decreasing to 0.029. These results demonstrate the effectiveness of our proposed modules in detecting camouflaged objects.

\noindent\textbf{Effectiveness of the Boundary-Aware Training Strategy.}
As described in Section 3.2, we propose a boundary-aware training strategy to enhance the boundary perception capability of the FEEM module. To investigate the usefulness of boundary expansion, we conduct an ablation study by applying dilation operations with different kernel sizes on the edge maps used in training. As shown in Table \ref{tab3}, we experiment with kernel sizes $ k \in \left \{ {1,3,5,7,9} \right \} $. 
When $k = 1$, the edge map remains in its original, unexpanded form.
The model achieves the best performance when $k=5$, demonstrating that appropriately expanded edge regions can serve as strong supervision signals, effectively guiding the model to produce more accurate boundary predictions.

%%%%%%%%%%%%%%%%%%%%%%%%%%%%%%%%%%%%%%%%%%%%%%%%%%%%%%%%%%%%%%%%%%%%%%%%

\section{Conlusion}
In this study, we propose a novel framework BASFNet, which introduces a boundary-aware spatial-frequency fusion strategy.
BASFNet effectively integrates frequency and spatial domain features, and the model can more accurately capture the boundary details of camouflaged objects and show significant robustness in complex backgrounds.
In addition, the combination of boundary-aware training strategies enables the model to better focus on camouflaged object boundaries, significantly improving the overall segmentation performance.
The experimental results demonstrate the superiority and effectiveness of BASFNet in COD tasks and highlight the effectiveness of fusing frequency and spatial information in COD tasks.

%%%%%%%%%%%%%%%%%%%%%%%%%%%%%%%%%%%%%%%%%%%%%%%%%%%%%%%%%%%%%%%%%%%%%%%%

%%% Use this environment to include acknowledgements (optional).
%%% This will be omitted in doubleblind mode.

% \begin{ack}
% By using the \texttt{ack} environment to insert your (optional) 
% acknowledgements, you can ensure that the text is suppressed whenever 
% you use the \texttt{doubleblind} option. In the final version, 
% acknowledgements may be included on the extra page intended for references.
% \end{ack}

%%%%%%%%%%%%%%%%%%%%%%%%%%%%%%%%%%%%%%%%%%%%%%%%%%%%%%%%%%%%%%%%%%%%%%%%

%%% Use this command to include your bibliography file.

\bibliography{mybibfile}

\end{document}